\pdfoutput=1 
\documentclass[10pt,twocolumn,letterpaper]{article}

\usepackage{iccv}
\usepackage{times}
\usepackage{graphicx}
\usepackage{amsmath}
\usepackage{amssymb}
\usepackage{graphicx}
\usepackage{booktabs}
\usepackage{multirow}
\usepackage{subcaption}
\usepackage[algo2e,ruled,vlined]{algorithm2e}
\usepackage{algorithm}
\usepackage{stfloats}
\usepackage{acronym}

\newcommand{\cutparagraphup}{\vspace*{-0.1in}}

\usepackage[pagebackref=true,breaklinks=true,letterpaper=true,colorlinks,bookmarks=false]{hyperref}

\iccvfinalcopy 

\def\iccvPaperID{9186} 

\ificcvfinal\fi

\begin{document}

\title{Towards Fast, Memory-based and Data-Efficient Vision-Language Policy}

\acrodef{vla}[VLA]{Vision-Language-Action}
\acrodef{llms}[LLMs]{Large Language Models}
\acrodef{vlms}[VLMs]{Vision Language Models}

\author{
  Haoxuan Li,
  Sixu Yan,
  Yuhan Li,
  Xinggang Wang\thanks{Corresponding author} \vspace{3pt} \\
  \small School of Electronic Information and Communications, Huazhong University of Science and Technology \\
  \small \texttt{\{lihaoxuan,yansixu,yuhanli,xgwang\}@hust.edu.cn} \\
  \\
  {
  \centering
    \textcolor{blue}{\href{https://hustvl.github.io/LiteVLP/}{https://hustvl.github.io/LiteVLP/}} 
    }
}
\maketitle

\newcommand{\model}{\text{LiteVLP}\xspace}
\newcommand{\multi}{\text{LiteVLP-m}\xspace}
\newcommand{\single}{\text{LiteVLP-s}\xspace}

\begin{abstract}
\ac{vlms} pretrained on Internet-scale vision-language data have demonstrated the potential to transfer their knowledge to robotic learning. However, the existing paradigm encounters three critical challenges: (1) expensive inference cost resulting from large-scale model parameters, (2) frequent domain shifts caused by mismatched data modalities, and (3) limited capacity to handle past or future experiences.
In this work, we propose \model, a lightweight, memory-based, and general-purpose vision-language policy generation model. \model is built upon a pre-trained 1B-parameter VLM and fine-tuned on a tiny-scale and conversation-style robotic dataset. 
Through extensive experiments, we demonstrate that \model outperforms state-of-the-art vision-language policy on VIMA-Bench, with minimal training time. Furthermore, \model exhibits superior inference speed while maintaining exceptional high accuracy. In long-horizon manipulation tasks, \model also shows remarkable memory ability, outperforming the best-performing baseline model by 18.8\%. These results highlight \model as a promising model to integrating the intelligence of \ac{vlms} into robotic learning.

\looseness=-1
\end{abstract}
\section{Introduction}

\begin{figure}[t]
\centerline{\includegraphics[width=0.98\linewidth]{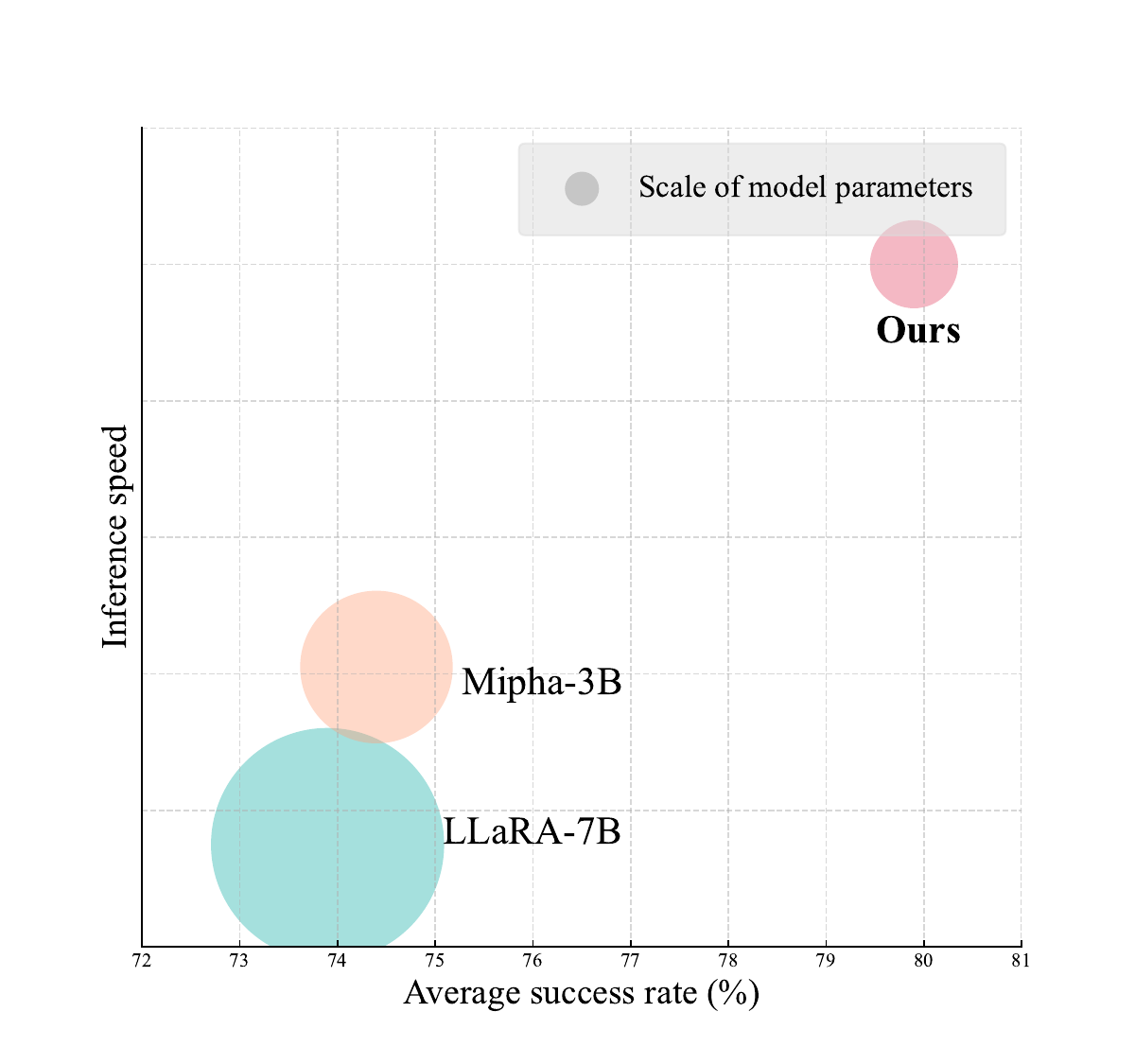}}
\caption{\textbf{Comparative performance of vision-language policies.} The x-axis represents the average success rate on VIMA-Bench, and the y-axis represents the inference speed evaluated on same devices. The bubble diameter indicates the number of model parameters}
\label{fig:success rate vs speed}
\end{figure}

Integrating pre-trained \ac{llms} and \ac{vlms} with low-level robotic policies enables context-aware robotic systems and enhances the robot’s ability to reason and interact with the environment~\cite{brohan2023rt2,o2023open,kim2024openvla,wen2024tinyvla,zheng2024tracevla,zhen20243d,cheng2024navila,cheang2024gr,zhou2025chatvla}.
Recently, the robotics domain has increasingly explored how to leverage \ac{llms} and \ac{vlms} for various robotic tasks such as perception, prediction, planning, and control~\cite{firoozi2023foundation}. 
However, despite the growing interest in these models, fully unlocking the capability of \ac{llms} and \ac{vlms} in robotics remains a great challenge.

Due to their remarkable performance in decision making and task reasoning, \ac{llms} and \ac{vlms} have been widely adopted for hierarchical task planning~\cite{ahn2022can,zeng2022socratic,driess2023palm,han2024interpret} and code generation~\cite{liang2023code,suris2023vipergpt,huang2023voxposer,duan2024manipulate}, facilitating guidance and interaction with low-level robotic controllers. To construct end-to-end models that generate robot actions directly from observations in natural vision and language, recent research has focused on developing \ac{vla} models~\cite{brohan2022rt1,brohan2023rt2,kim2024openvla,niu2024llarva,zheng2024tracevla,zawalski2024robotic,wen2024tinyvla,wen2024diffusion} that learn from web and robotics data. Some \ac{vla} models leverage pre-trained \ac{llms} and \ac{vlms}, which represent actions as simple text strings and tokenize them into text tokens with a tokenizer~\cite{brohan2022rt1,kim2024openvla,niu2024llarva}, while others utilize \ac{llms} and \ac{vlms} to compress multimodal representations and train additional action actors to refine action trajectory outputs~\cite{brohan2023rt2,black2410pi0,liu2024rdt,pertsch2025fast}.

However, there are three key challenges existing in the above-mentioned models. First, fine-tuning pre-trained \ac{llms} and \ac{vlms} with robotic data often encounters frequent domin shifts due to the substantial differences between the pre-training web dataset and the fine-tuning robotic dataset. Second, current models suffer from insufficient memory for future or past experiences, making them achieve poor performance for long-horizon manipulation. Third, the large number of model parameters in the backbone network leads to a high computational time, which limits their real-world deployment.

To solve the above challenges, in this paper, we introduce \model, a lightweight vision-language policy tailored to handle memory-dependent robotic tasks with a stable fine-tuning strategy that effectively leverages pre-trained knowledge. \model fine-tunes a pre-trained VLM backbone on a conversation-style robotic dataset via visuomotor instruction tuning training paradigm~\cite{li2024llara}. Due to the high structural alignment between robotic data and vision-language pre-training data, \model effectively mitigates frequent domin shifts during fine-tuning, resulting in greater training stability compared to traditional \ac{vla} models. To address long-horizon and memory-dependent tasks, \model supports multi-image input to better leverage past and future experiences. Meanwhile, \model introduces the RLT~\cite{choudhury2025don} in video transformers~\cite{arnab2021vivit, bertasius2021space, li2022mvitv2, fan2021multiscale} and modifies it as a multi-observation compression (MOC) module in our model to reduce the spatio-temporal complexity of the model. As a result, the number of image tokens decreases by approximately 60\%, while preserving the representational quality of the image patches. Furthermore, \model uses InternVL2-1B~\cite{Chen2024HowFA} as its backbone, and the small parameters of the model make its training and inference more efficient compared to previous models.
All of these characteristics make \model not only demonstrate computational efficiency, but also have great potential to improve real-time performance and scalability in long-horizon robotic manipulation tasks.

We conduct extensive experiments and demonstrate that \model can successfully handle a variety of robot manipulation tasks by fine-tuning a lightweight VLM on a small robotic manipulation dataset. It achieves competitive results on the VIMA-Bench~\cite{jiang2023vima}. In general, our contributions can be summarized as follows: \looseness=-1
\begin{itemize}
    \item We propose \model, the first vision-language policy with efficient inference, memory effects, and fine-tuning stability. It employs a lightweight VLM as the backbone, supports future goals or past experiences as input, and fine-tunes with conversation-style training manner.
    \item We introduce a novel MOC module in \model to improve training efficiency and accelerate inference speed. This module effectively reduces the number of image tokens, increasing inference speed by 34.1\%.
    \item We also conduct extensive experiments on 17 simulated tasks from VIMA-Bench. With equitable training on the same small dataset, \model outperforms the state-of-the-art vision-language policy while delivering a 70\% reduction of training time and 6.8 times acceleration of inference speed. Furthermore, in long-horizon tasks, \model achieves a 18.8\% higher success rate compared to baselines.
\end{itemize}
 
\section{Related Work}

\noindent\textbf{\ac{llms} and \ac{vlms} in Robot Learning}
Recently, \ac{vlms} have made significant progress in various tasks. Models such as InternVL~\cite{Chen2024HowFA}, Qwen2-VL~\cite{Wang2024Qwen2VLEV}, Flamingo~\cite{alayrac2022flamingo}, BLIP-2~\cite{Li2023BLIP2BL}, Mipha~\cite{zhu2024mipha}
and LLaVA~\cite{liu2023visual} have demonstrated their remarkable potential and application in multimodal learning tasks. 
With the rise of embodied intelligence, integrating \ac{llms} and \ac{vlms} into robotics has become a promising research direction, catalyzing the development of both generalist robot policies~\cite{brohan2022rt,brohan2023rt,team2024octo,kim2024openvla,black2410pi0,liu2024rdt} 
and \ac{vla} models~\cite{o2024open,kim2024openvla,zawalski2024robotic,zheng2024tracevla,wen2024diffusion,li2023vision,nair2022r3m,shridhar2022cliport,huang2023embodied}. 
However, existing generalist robot policies or \ac{vla} models typically have a large number of parameters, as they are fine-tuned from large \ac{vlms}, resulting in high training and deployment costs. 
The method proposed in this paper demonstrates that fine-tuning a small-parameter VLM can achieve comparable performance to large-scale generalist robot policies while significantly reducing computational costs and improving inference efficiency.

\smallskip

\noindent\textbf{Memory-based Robotic Manipulation}
In robotics domain, enabling robots to think human-like is a long-standing and unsolved challenge. Therein, equipping robots with memory capabilities is particularly crucial, 
as it allows robots to store and retrieve historical experiences, thus enhancing their ability to perform complex tasks efficiently~\cite{jockel2008towards,vernon2022cognitive}.
Early studies on robotic memory-based manipulation mainly addressed navigation using constrained semantic maps~\cite{bowman2017probabilistic,chaplot2020object}, 
while other studies focused on designing memory models and representations within cognitive architectures. Recent advancements like SAM2Act+~\cite{fang2025sam2act} inspired by SAM2~\cite{ravi2024sam}, incorporate a memory bank, an encoder, and an attention mechanism to improve spatial memory for robotic tasks. A slight difference between the above methods and our work is that \model not only integrates the memory of historical experiences but also future goal states, which enables the robot to handle long-horizon tasks more effectively.
\smallskip

\noindent\textbf{Robotic Manipulation}
Recently, some studies have attempted to integrate visual observations with robotic proprioception to achieve precise predictions for robotic manipulation, such as Where2Act~\cite{mo2021where2act}, Flowbot3d~\cite{eisner2022flowbot3d}, Partnet~\cite{mo2019partnet} and Sparsedff~\cite{wang2023sparsedff}. 
With the rapid advancement of multimodal large language models (MLLMs)~\cite{awadalla2023openflamingo,li2023blip,Li2023BLIP2BL,liu2023visual}, their powerful reasoning capabilities have introduced new ways to solve robotic manipulation. OpenVLA~\cite{kim2024openvla} is an open-source \ac{vla} model that integrates Llama2 ~\cite{touvron2023llama} with DINOv2~\cite{oquab2023dinov2} and SigLIP~\cite{zhai2023sigmoid}, achieving strong generalist manipulation capabilities. 3D-VLA~\cite{zhen20243d} introduces a new family of embodied foundation models based on a 3D LLM~\cite{hong20233d}, 
seamlessly linking 3D perception, reasoning, and action through a generative world model. $\pi_0$~\cite{black2410pi0} is a novel architecture based on a pre-trained VLM and a flow matching~\cite{liu2022rectified} action expert, 
enabling robots to achieve strong generalization capabilities and execute complex and highly dexterous tasks. However, these models typically rely on large-parameter VLM backbones 
and require training on extremely large-scale datasets. In contrast, our model utilizes only a 1B-parameter pre-trained VLM, achieving comparable or even superior performance to large-scale models while maintaining high data efficiency.

\section{Method}

\begin{figure*}[t!]
\centerline{\includegraphics[width=0.98\linewidth]{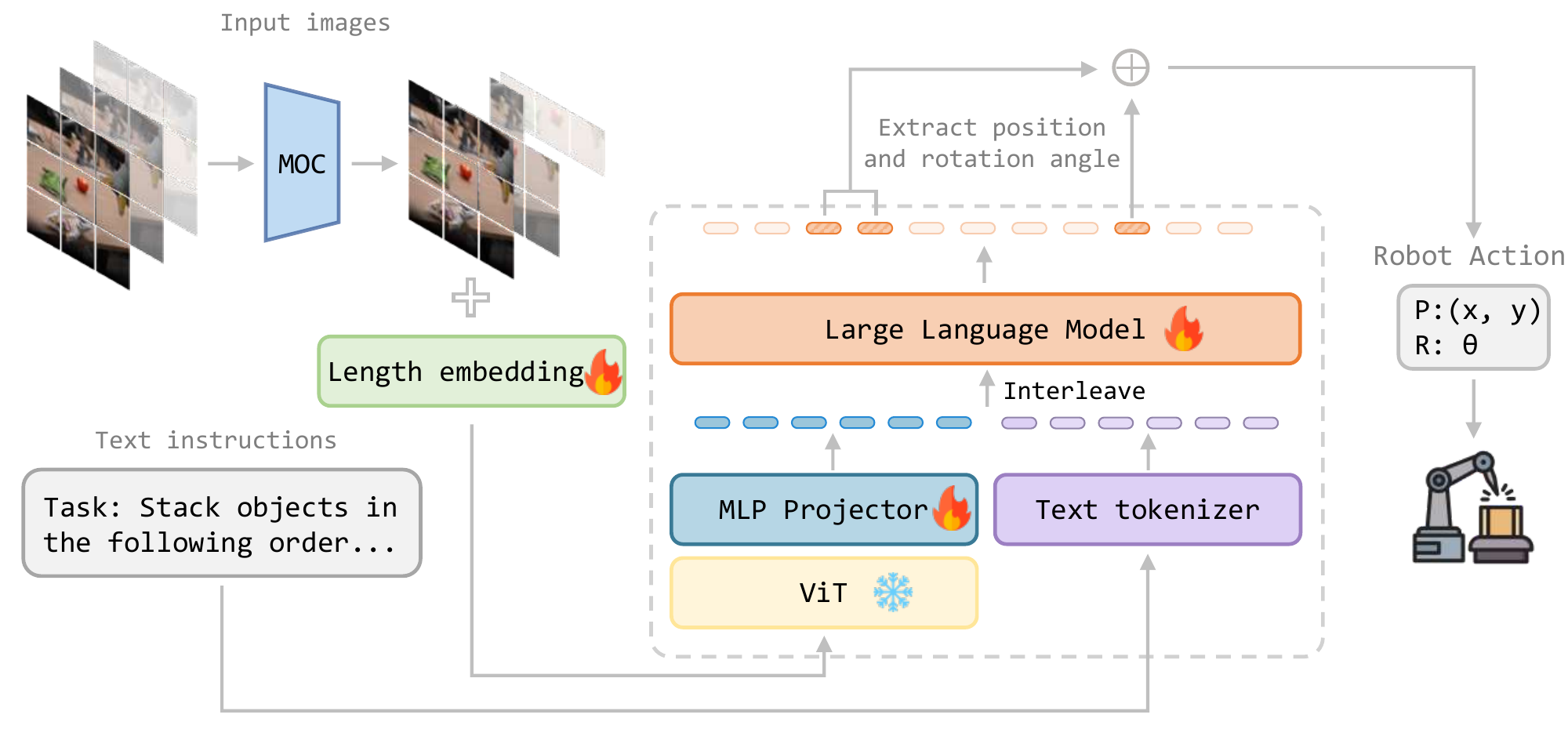}}
\caption{\textbf{Overall framework of \model.} The \model initiates with multi-observation compression and then projects the image features into the same dimensional space as the text features. Subsequently, the image tokens are interleaved with text tokens and processed by a large language model to generate a text output that includes the end-effector's action. Of note, during the fine-tuning stage, the parameters of the ViT are frozen, while the length embedding, the MLP projector and the large language model are trained.}
\label{fig:overall framework}
\end{figure*}

\subsection{Problem Formulation}  
The problem of robot manipulation conditioned on vision-language can be formally modeled as a structured decision-making task that generates the current action of the robot based on the visual observations, task instructions, and future goal images given. Let $\mathcal{M} = \left\langle \mathcal{S}, \mathcal{O}, \mathcal{G}, \mathcal{A} \right\rangle$ represent the overall problem framework. Here:
\begin{itemize}
    \item $\mathcal{S}$ is the state space, where a state $s_t \in \mathcal{S}$ represents the configuration of the robot and the environment at timestamp $t$.   
    \item $\mathcal{O}$ is the observation space, where $o_t \in \mathcal{O}$ denotes the observation at timestep $t$, consisting of both visual and textual information about the environment.  
    \item $\mathcal{G}$ is the goal condition, which includes a set of goal images and textual instructions that define the desired outcome of the task.    
    \item $\mathcal{A}$ is the action space, where an action $a_t \in \mathcal{A}$ represents the robot action at timestamp $t$.
\end{itemize}  

In the context of the generation of vision-language policy, the goal is to generate a sequence of actions that enable the robot to perform the task specified by the goal condition $\mathcal{G}$. Our model \model can be defined as a policy function $\pi$ that maps the current state $s_t$, visual observation $o_t$ and goal condition $g_t$ to an action $a_t$, described as: 
\begin{equation}
a_t = \pi(s_t, o_t, g_t; \theta)
\end{equation}
where $\theta$ represents the parameters of the policy model. After executing each action, the new observation $o_{t+1}$ is obtained. The state $s_{t+1}$ is updated by executing the action $a_t$ and the state $s_t$. The process continues iteratively until the goal condition $\mathcal{G}$ is satisfied or a termination condition is met.  

\subsection{Model architecture}
In this work, our goal is to develop an end-to-end vision-language robot policy that fully utilizes the strong semantic reasoning ability of \ac{vlms} to instruct robotic manipulation. The architecture of our model is illustrated in Fig.~\ref{fig:overall framework}. At timestamp $t$, our model $\pi$ accepts multiple images $I_{N,t} \in R^{N \times W \times H \times 3}$ and language instruction $L_t$ as input, finally generating a pure language answer $L_{a,t}$ that contains the robot's impending actions. 
Here, $N$ is the number of images, $W$ and $H$ are the width and height of the image respectively.
Before $I_{N,t}$ are inserted into the corresponding positions in the text, they first undergo the MOC module. After an extraction process $Ext$, the 2D image coordinates $C_{i,t}$ and the rotation angle $R_{i,t}$ of the end effector can be extracted from answer $L_{a,t}$, which define a robot's action $A_t(C, R)$.  
Meanwhile, the 2D image coordinates undergo a transformation into the robot's action space through a predefined mapping process, 
which can be determined via visual calibration in both simulated and real-world settings. 
\par
Once the above round is completed, the robot performs the specified action, capturing a new observation for subsequent processing. 
The entire process can be represented by the following formulation: \looseness=-1
\begin{equation}
    L_{a,t} = \pi\left(\text{MOC}\left(I_{N,t}\right), L_t\right)
\end{equation}
\begin{equation}
    A_t(C, R) = Ext\left(L_{a,t}\right)
\end{equation}
where $A_t(C, R)=\left\{ \mathbf{q}_{i,t} = (x_{i,t}, y_{i,t}, \theta_{i,t}) \mid i = 1, \dots, N_A \right\}$. Meanwhile, $C_{i,t}=(x_{i,t}, y_{i,t})$, $R_{i,t}=\theta_{i,t}$, 
$N_{A,t}$ represents the number of actions generated.

\subsection{Memory-based Visuomotor Instruction Tuning}
We adopt the visuomotor instruction tuning method proposed in LLaRA~\cite{li2024llara} to fine-tune \model, leveraging its effectiveness in addressing distribution shifts when fine-tuning pre-trained VLMs with robotic data. In multi-step tasks, previous actions $A_{1:t-1}$ and their outcomes or historical observations $O_{1:t-1}$ 
are critical to generate subsequent actions. Notably, $I_N$ already consists of $O_{1:t-1}$. 
Otherwise, we explicitly append $A_{1:t-1}$ to $L_t$ in the form like "You have finished ...", enabling the model to obtain the whole process of the task. 
This integration of historical information preserves the model's memory for long-horizon tasks, enhancing its decision-making capability in sequential actions.\looseness=-1

\subsection{Multi-observations compression}

\begin{figure}[H]
\centerline{\includegraphics[width=0.99\linewidth]{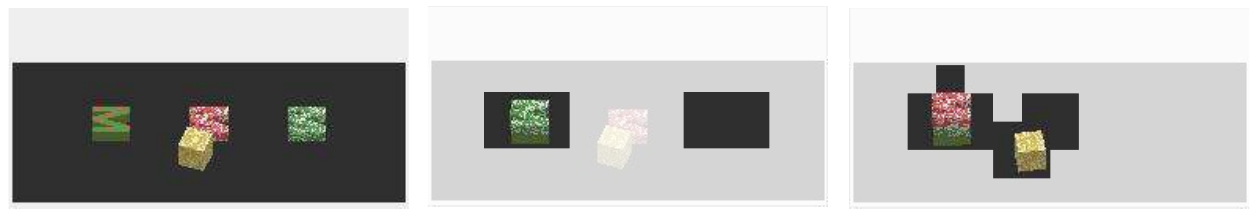}}
\caption{\textbf{Simple visualization of MOC's effect.} The light gray image patches indicate unchanged areas between consecutive images, which will be reduced in the sequence of image patches.}
\label{fig: MOC_method}
\end{figure}

We observe a common characteristic in robot manipulation tasks: there is a significant amount of redundant image information between multiple consecutive observation images. 
Our goal is to identify whether patches at the same spatial location in two consecutive images are static. 
These static patches correspond to background elements or objects in the images that do not change over time. 
The results after the removal of static patches are shown in Fig.~\ref{fig: MOC_method}. To optimize efficiency, we replace subsequent static patches with the patch that first appeared. 
Consequently, each compressed patch requires a length embedding to indicate how many images it persists across.
The process of implementing the above method is as follows: First, we compare image patches at the same spatial location across consecutive images and set a threshold $\epsilon$. If this condition is met, we consider $P_{x, y}^i$ to be static: 
\begin{equation}
    \left\|P_{x, y}^{i+1}-P_{x, y}^i\right\|<\epsilon
\end{equation}
Thus, we can obtain an image patch mask $M_{S}^N$ to indicate static patches and a run-length mask $M_{L}^N$ to represent 
the durative length of each static patch. The multi-image compression process can be formulated as follows:
\begin{equation}
    \text{MOC}\left(I_{N,t}\right)=M_{S}^N\ \, \circ \, P\left(I_{N,t}\right)+M_{L}^N \, \odot \, Emb_{L}
\end{equation}
Here, $P$ represents the operation of performing position embedding on the image and $Emb_{L}$ is a learnable length embedding parameter.

\section{Experiments}

\subsection{Experimental Setup}
\paragraph{Benchmark}
VIMA-Bench~\cite{jiang2023vima} is a simulated tabletop robotic manipulation environment for evaluating \ac{vlms} 
using multimodal instructions. It includes 17 tasks, each with text and reference images guiding the robot's actions, 
such as \texttt{follow order} and \texttt{rearrange}. The robot's action space consists of 2D coordinates for placement and quaternions for rotations. 
VIMA-Bench uses a four-level protocol to assess generalization: placement (L1), combinatorial (L2), novel object (L3), and novel task generalization (L4). 
This setup provides a comprehensive evaluation of \ac{vlms} in robotic manipulation.  \looseness=-1

\begin{figure*}[t!]
\centerline{\includegraphics[width=0.95\linewidth]{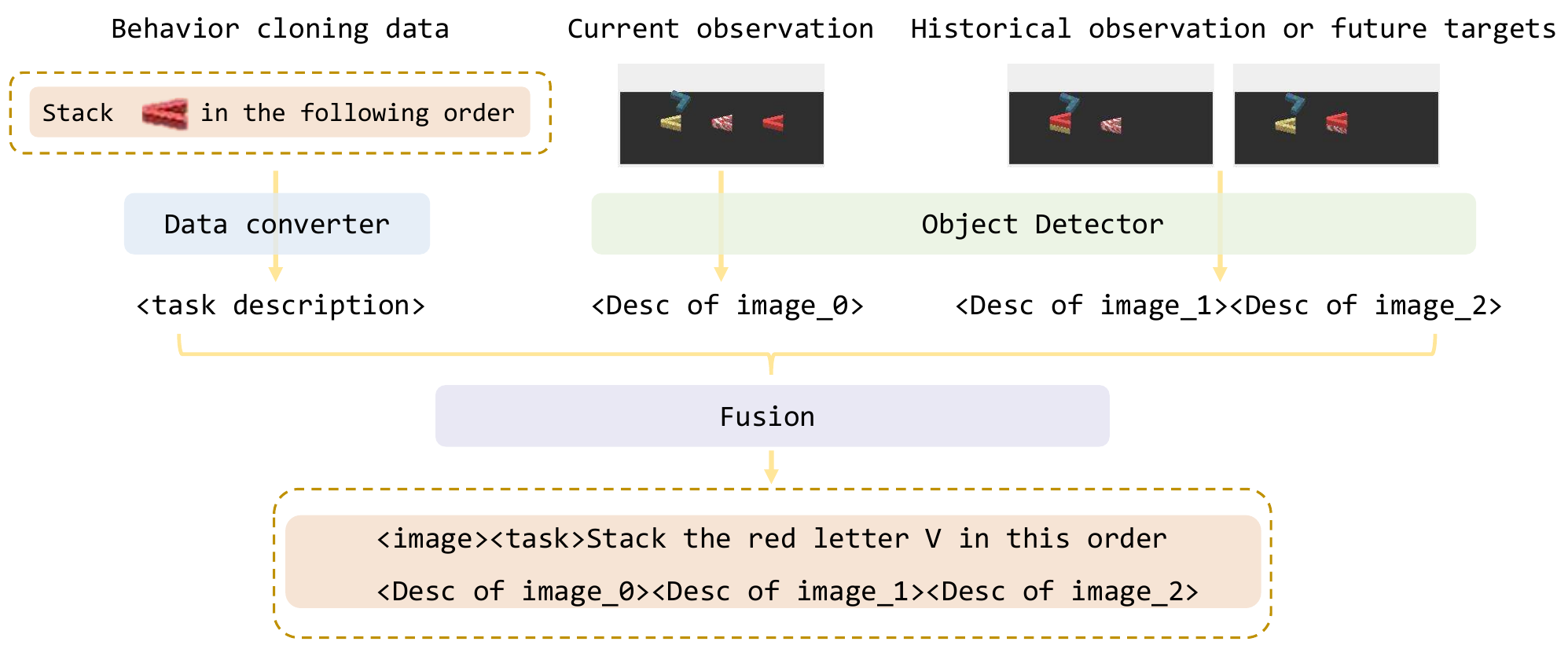}}
\caption{\textbf{Example of D-inBC dataset format.} The D-inBC dataset includes the task description and the description of reference images. A data converter processes instructions, while an object detector extracts the locations of each object to form the image descriptions.}
\label{fig:LLaRA_dataset}
\end{figure*}

\cutparagraphup
\paragraph{Training datasets}
We use the dataset introduced in LLaRA, which provides three datasets of varying sizes. For our training, we select the Description-Instruct-BC-8k (D-inBC-8k) dataset. We present an example of D-inBC in Fig.~\ref{fig:LLaRA_dataset}. Note that the inputs of \multi and \single
are diverse. The input of \multi includes complete images and the corresponding object detection results. 
The complete images comprise historical observations, current observations, and future goal images. However, the input of \single only includes current observation. And the way \single obtains the information of complete images is through the object detection results.  \looseness=-1

\cutparagraphup
\paragraph{Evaluation tasks}
VIMA-Bench provides tasks with difficulty levels ranging from L1 to L4. We select L1 to L3 as test tasks to evaluate the performance of our model. The reason we abandon the L4 task is the mismatch of spatula's rotation data between training and testing phases. This inconsistency can significantly affect the success rate of the L4 task, so we decided to only use L1 to L3 as our test tasks~\cite{li2024llara}.  \looseness=-1

\cutparagraphup
\paragraph{Implementation details}
During the fine-tuning stage, we freeze the vision encoder parameters and train the MLP layers, the large language model, and the newly introduced length embedding layers in MOC module. The hyperparameters are listed in Tab.~\ref{Tab:hyperparameter}. We train \model on D-inBC-8k for 4 epochs. We use the AdamW optimizer with $(\beta_1,\beta_2)=(0.9,0.999)$, a learning rate of 2e-5, a weight decay of 0.01, and a warm-up ratio of 0.03. The learning rate schedule is CosineAnnealingLR. Furthermore, in the MOC module, we set the threshold $\epsilon$ to 1e-5 during the fine-tuning and inference stages to maximize the retention of informative image patches from observations.
\begin{table}[!h]
\centering
\scalebox{0.99}{
\begin{tabular}{c c}
    \toprule
    Hyperparameter & Value \\ \midrule
    learning rate & 2e-5 \\
    epochs & 4  \\
    optimizer & AdamW \\
    weight decay & 0.01 \\
    warming up & 2e-5 \\
    lr schedule & CosineAnnealingLR \\
    $\epsilon$ & 1e-5 \\\bottomrule
    \end{tabular}}
\caption{\textbf{Hyperparameters used during fine-tuning.} This table presents the hyperparameters during model fine-tuning. Here, the parameter $\epsilon$ represents the threshold in the MOC module.}
\label{Tab:hyperparameter}
\end{table}

\subsection{Evaluation Results and Analysis}

\begin{figure*}[!h]
\centerline{\includegraphics[width=0.87\linewidth]{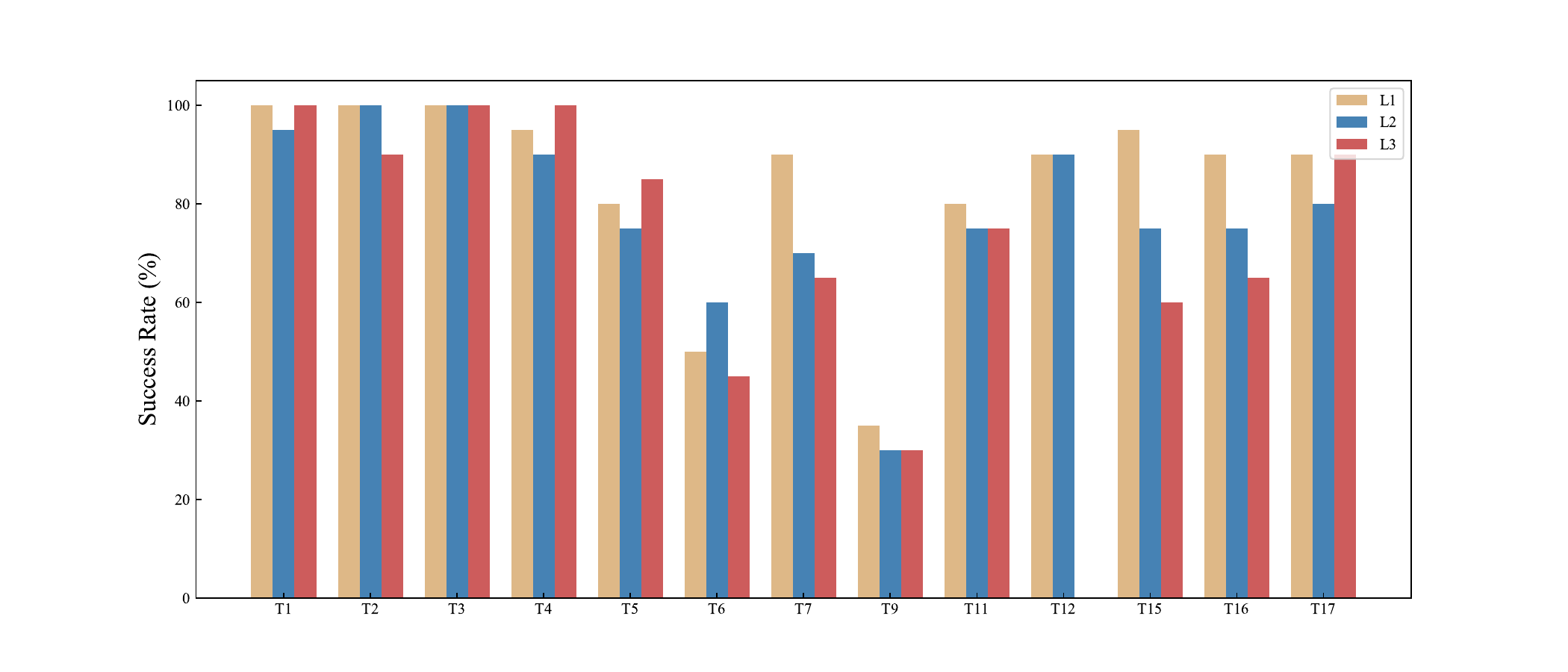}}
\caption{\textbf{Success rates of all tasks.} Note that the difficulty level L3 doesn't include \texttt{sweep without exceeding} task.}
\label{fig:total success}
\end{figure*}

\paragraph{Evaluation results} 
We evaluate our method on VIMA-Bench and compare its performance against other models trained on datasets of different sizes. 
The results are shown in Tab.~\ref{Tab:baseline}.  We demonstrate that \model achieves highly competitive performance, achieving a success rate of 84.2\% on L1, 78.1\% on L2, and 75.4\% on L3, using only 1.2\% of the VIMA dataset. This performance is comparable to the state-of-the-art model VIMA, which achieves 81.5\% on L1, 81.5\% on L2, and 78.7\% on L3 when trained on the full dataset. Additionally, our model significantly outperforms other \ac{vlms} fine-tuned on our small dataset, such as LLaRA-7B and Mipha-3B. These results successfully indicate that \model can rapidly adapt to robotic manipulation and demonstrate highly competitive performance when fine-tuned with visuomotor instruction in a small robotic dataset. The success rates of all tasks at three difficulty levels are shown in Fig.~\ref{fig:total success}.
\begin{table}[!h]
\centering
\scalebox{0.94}{
\begin{tabular}{l|c|ccc}
    \toprule
    Model & Data & L1 (\%) & L2 (\%) & L3 (\%)
    \\ \midrule
    VIMA & 100\% & 81.5 & \textbf{81.5} & \textbf{78.7} \\
    VIMA & 10\% & 76.3 & 75.8 & 73.2 \\
    VIMA & 1\% & 36.3 & 34.3 & 15.4 \\
    LLaRA-7B-m & 1.2\% & 44.6 & 27.7 & 34.2 \\
    LLaRA-7B-s & 1.2\% & 78.1 & 73.8 & 70.0 \\
    Mipha-3B & 1.2\% & 78.8 & 72.3 & 72.1 \\\midrule
    \single & 1.2\% & 83.1 & \textbf{81.5} & 77.5 \\
    \multi & 1.2\% & \textbf{84.2} & 78.1 & 75.4 \\\bottomrule
    \end{tabular}}
\caption{\textbf{Evaluation results on VIMA-Bench.} Comparison of success rates over different models trained on datasets of varying sizes. For brevity, the suffix '-m' in model names denotes multi-image input, while '-s' indicates single-image input.}
\label{Tab:baseline}
\end{table}

\begin{figure}[!h]
\centerline{\includegraphics[width=0.98\linewidth]{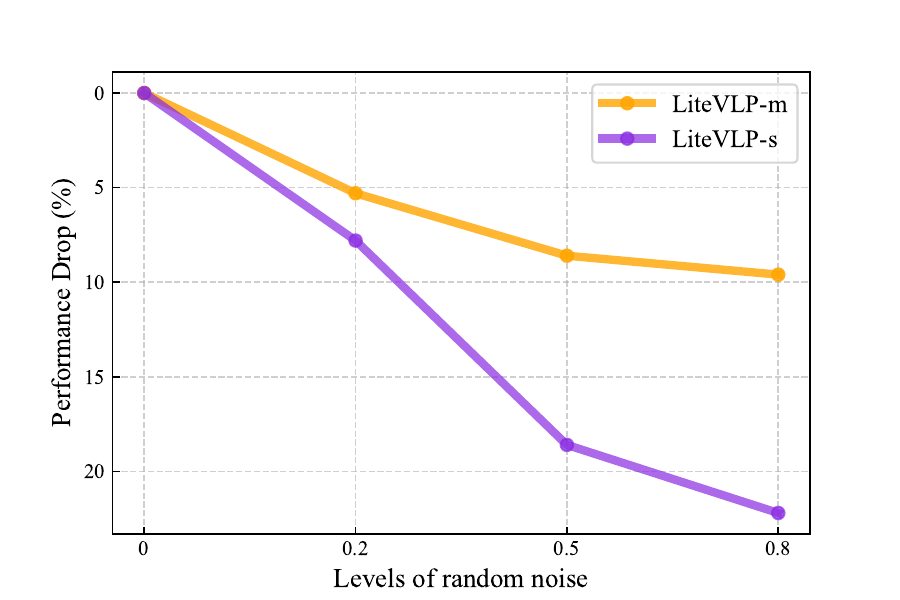}}
\caption{\textbf{Robustness performance.} The figure illustrates the performance drop under noise levels of 0.2, 0.5, and 0.8.}
\label{fig:noise}
\end{figure}

\cutparagraphup
\paragraph{Robustness to object location}
We simulate the scenario of inaccurate object localization in real-world applications to test the model's performance stability, 
with the results shown in Fig.~\ref{fig:noise}. By adding random noise of varying sizes to the object location coordinates, 
we assess the average decline in task success rate. As shown in the table, with a noise level of 0.2, \multi drops by 5.3\% and \single by 7.8\%. 
And when the noise level increases to 0.8, the drops of \multi and \single rise to 9.6\% and 22.2\% each. 
Although the average performance drops for both models as the noise level increases, compared to \single, 
\multi drops much smaller in task success rate. 
This is because \single can only obtain object location information from the object detection text description, 
while \multi can obtain location information both from the text description and the image data across multiple views, 
effectively enhancing the model's robustness to location inaccuracies.

\cutparagraphup
\paragraph{Long-horizon manipulation performance}
In long-horizon manipulation tasks, \multi significantly outperforms other baseline models , the results are shown in Tab~\ref{Tab:long_horizon}. 
We refer to CoTDiffusion~\cite{ni2024generate} to select three representative long-horizon manipulation tasks in VIMA-Bench \textemdash  \texttt{visual rearrangement}, \texttt{visual reasoning}, 
and \texttt{visual constraints}. CoTDiffusion is a model specifically designed to improve performance in long-horizon manipulation tasks. However, 
our \multi demonstrates superior performance in long-horizon manipulation tasks, achieving an average improvement of 18.8\% over CoTDiffusion in three types of tasks. The effectiveness of our model can be attributed to the sufficient memory for past and future experiences, making it more capable of long-horizon 
manipulation.

\begin{table}[!h]
\centering
\scalebox{0.90}{
\begin{tabular}{lcccc}
    \toprule
    \multirow{2}{*}{Model} & Rearrange & Reasoning & Constraints & Avg \\
          & (\%) & (\%) & (\%) & (\%)
    \\ \midrule
    Gato & 6.4 & 2.5 & 25.2 & 11.4 \\
    Flamingo & 17.5 & 3.0 & 36.1 & 18.9 \\
    SuSIE & 37.7 & 39.0 & 52.3 & 43.0 \\
    VIMA & 43.1 & 38.2 & 67.2 & 49.5 \\
    CoTDiffusion & 59.0 & 51.7 & 83.1 & 64.6 \\\midrule
    \multi & \textbf{80.0} & \textbf{86.7} & \textbf{83.4} & \textbf{83.4} \\\bottomrule
    \end{tabular}}
\caption{\textbf{The evaluations on three typical long-horizon tasks.} We refer to CoTDiffusion to select three representative long-horizon manipulation tasks in VIMA-Bench.}
\label{Tab:long_horizon}
\end{table}

\subsection{Efficiency Analysis}

In this section, we evaluate the efficiency of our method in both training and inference. 
Our approach strikes a balance between performance and computational cost, achieving competitive accuracy with significantly reduced training time and faster inference speed.

\paragraph{Analysis of training time}
Based on the results present in Tab.~\ref{Tab:training time}, our method demonstrates a significant advantage on training efficiency. Specifically, we achieve an average success rate of 80.7\% in just 6.1 hours of training, using 4 NVIDIA RTX 3090 GPUs. In comparison, VIMA, which is trained on 8 NVIDIA V100 GPUs, takes 24 hours and achieves an average success rate of 80.6\%, while LLaRA-7B, trained on 4 NVIDIA RTX 3090 GPUs, requires 21 hours and achieves 74\% on average. These results highlight the efficiency of our approach, which not only reduces training time significantly by 17.9 hours compared to VIMA and 14.9 hours compared to LLaRA-7B but also performs excellently even on less powerful GPU setups.

\begin{table}[!h]
    \centering
    \scalebox{0.85}{
    \begin{tabular}{l|c|c|c}
        \toprule
        Model & Training time & Device (\%) & Avg (\%) 
        \\ \midrule
        VIMA & 24h & 8*NVIDIA V100 & 80.6 \\
        LLaRA-7B & 21h & 4*NVIDIA RTX 3090 & 74.0 \\
        LiteVLP & \textbf{6.1h} & 4*NVIDIA RTX 3090 & \textbf{80.7} \\\bottomrule
        \end{tabular}}
    \caption{\textbf{Training time when achieving the highest success rate.} Comparison of training time over different models trained on 
    GPUs with different performance levels.}
    \label{Tab:training time}    
\end{table}

\cutparagraphup
\paragraph{Fast inference speed}
With a lightweight design, our model not only significantly reduces training time, but also accelerates inference speed, demonstrating a huge advantage on low latency. We fairly compare our \multi with Mipha-3B and LLaRA on VIMA-Bench tasks, with the same NVIDIA RTX 3090. As shown in fig~\ref{fig:success rate vs speed}, 
our \multi achieves a superior performance with 6.8 times lower inference latency than LLaRA. 
This result can be attributed to two main factors. First, our model contains only 1B parameters, 
which is smaller than the 7B parameters of LLaRA, greatly reducing computational overhead. 
Second, our method of multi-observation compression effectively reduces the number of image tokens, thereby shortening the input sequence length and thus accelerating inference speed. Fig.~\ref{fig:MOC} shows four visualization examples of images processed by the MOC modeule. 

\subsection{Ablation Studies}
In this section, we conduct several additional experiments to research the effect of multi-observation compression and the position of multiple image tokens.

\begin{figure}[H]
\centerline{\includegraphics[width=1\linewidth]{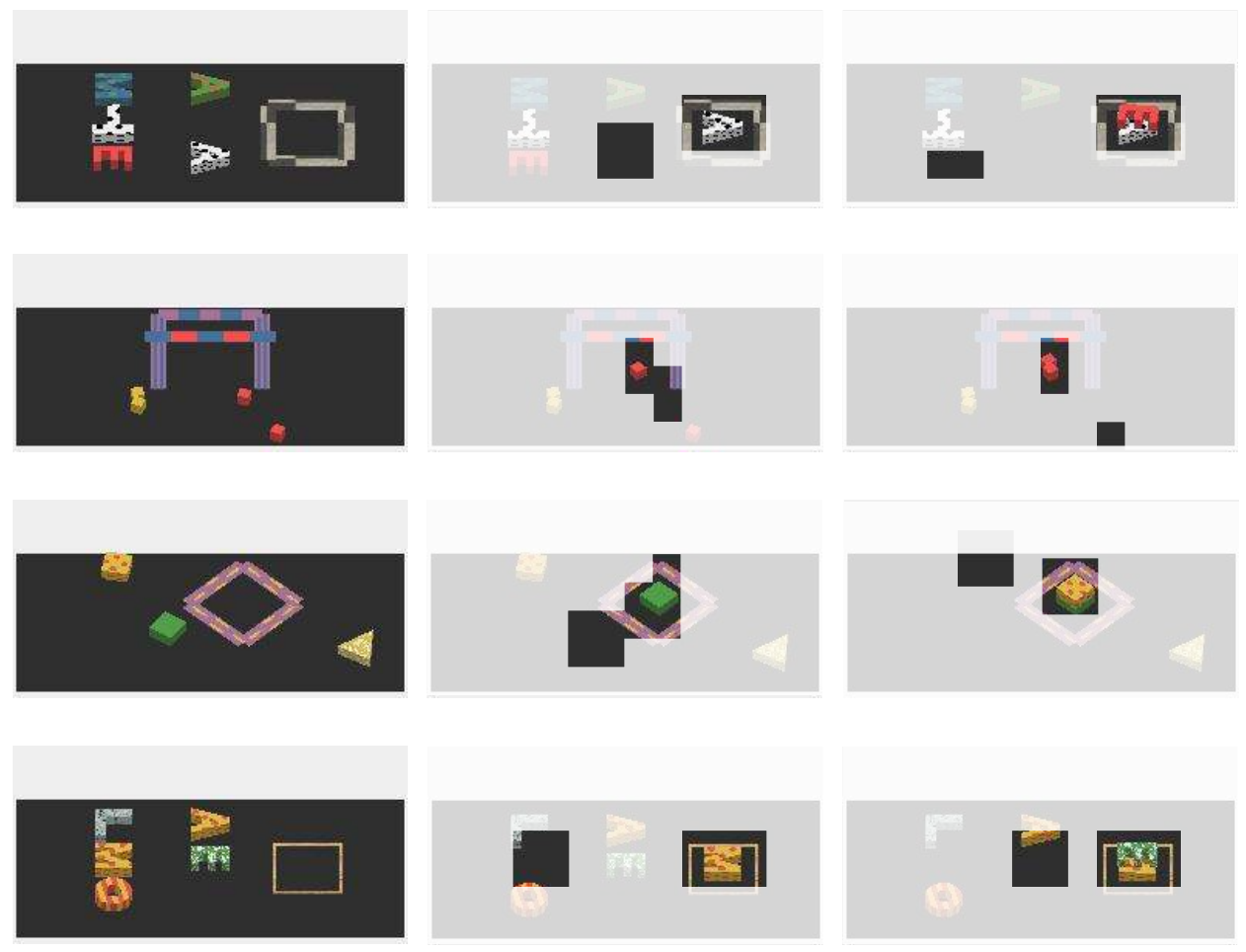}}
\caption{\textbf{Visualization of MOC results.} Here we present more visualization results of using MOC in different manipulations.}
\label{fig:MOC}
\end{figure}

\paragraph{The effect of multi-observation compression}
We analyze the impact of multi-observation compression on training time and inference speed by comparing its use versus non-use. The results are shown in Tab~\ref{Tab:ablation_MOC}. We observe that adopting multi-observation compression has minimal impact on task success rate but significantly reduces training time by 47.5\% and increases inference speed by 34.1\%. This can be explained by the change in input sequence length. In multi-image input data, image tokens constitute the majority of the sequence. The visualization results of the MOC module are shown in Fig.~\ref{fig:MOC}. The MOC module effectively reduces the number of these tokens, leading to a shorter sequence length, which in turn decreases training time and enhances inference speed.

\begin{table}[!h]
    \centering
    \scalebox{0.91}{
    \begin{tabular}{l|ccc|c}
        \toprule
        \multicolumn{1}{c|}{Model} & L1(\%) & L2(\%) & L3(\%) & avg(\%) \\ \midrule
        \model (w/ MOC) & 84.2 & \textbf{78.1} & 75.4 & 79.2  \\
        \model (w/o MOC) & \textbf{84.6} & 75.0 & \textbf{79.3} & \textbf{79.6} \\
        \bottomrule
    \end{tabular}}
    \caption{\textbf{Ablation studies on the use of multi-observation compression.} This table illustrates the comparative analysis of task success rates with and without the activation of the MOC module.}
    \label{Tab:ablation_MOC}
    \end{table}

\cutparagraphup
\paragraph{Position of multiple image tokens}
Tab~\ref{Tab:ablation_position} illustrates the effect of the positions of multiple image tokens. Meanwhile, "Collection" means that all image tokens are gathered at the beginning of the prompt, while "Interleaved" indicates that image tokens are dispersed throughout the text. From the result, we can easily know that compared to "Collection", the "Interleaved" method results in a significantly lower average success rate of 71.2\%, representing a 10.1\% performance drop. The performance gap suggests that aggregating image tokens at the beginning provides a more structured input representation, allowing the model to process visual information more effectively. Conversely, interleaving image tokens within the text may introduce discontinuities that hinder the model's ability to integrate multimodal information efficiently. These findings indicate that positioning image tokens at the beginning of the prompt is a more effective strategy to improve task success rates.

\begin{table}[!h]
\centering
\scalebox{0.96}{
\begin{tabular}{c|ccc|c}
    \toprule
    Position & L1(\%) & L2(\%) & L3(\%) & avg(\%)\\ \midrule
    Collection & \textbf{84.2} & \textbf{78.1} & \textbf{75.4} & \textbf{79.2}\\
    Interleaved & 72.7 & 69.6 & 71.2 & 71.2(10.1\%$\downarrow$) \\\bottomrule
    \end{tabular}}
\caption{\textbf{Ablation studies on the position of multiple image tokens.} Position represents the positional relationship of image tokens inserted among text tokens. This table presents a comparison of task success rates under different insertion positional relationships.}
\label{Tab:ablation_position}
\end{table}

\section{Limitations and Future Work}
\subsection{Limitations}
The primary limitations of \model include two aspects: its inability to operate effectively in 3D environments and its difficulty in handling contact-rich manipulation tasks. In the following, we discuss these challenges in detail.
\looseness=-1

\cutparagraphup
\paragraph{Limitations in 3D robotic manipulations}
Since our model relies solely on 2D RGB images as input, it inherently lacks depth perception, which is crucial for understanding spatial structures in 3D environments. Additionally, because its capability is partially constrained by the coordinates of 2D object detection to identify target objects, extending its applicability to 3D scenes poses significant challenges. The absence of depth cues makes it difficult for the model to generalize beyond planar understanding, limiting its effectiveness in scenarios that require full spatial awareness. 
\looseness=-1

\cutparagraphup
\paragraph{Incapability of completing contact-rich manipulation tasks} The output of our model includes two parts: the position and the rotation of the end effector. However, contact-rich manipulation tasks, such as assembling components, inserting objects, or handling deformable materials, require fine-grained force control and real-time adaptation to unpredictable physical interactions. Since our model relies primarily on visual and text input without explicit force or tactile feedback, it lacks the necessary sensory information to regulate interaction forces effectively. In addition, precise manipulation often requires compliance control, friction estimation, and dynamic adaptation, which are challenging to achieve with purely vision-language-conditioned policies.

\subsection{Future work}
In future work, we aim to enhance the model’s performance in 3D environments and contact-rich manipulation. Integrating depth perception from depth sensors improves the spatial understanding, while fusing 2D and 3D data with self-supervised or contrastive learning may enhance reasoning and generalization. For contact-rich tasks, multi-modal feedback from force-torque and tactile sensors can enable adaptive interaction. Reinforcement learning in physics-based simulations may refine compliance control and dexterous manipulation. Additionally, hybrid policies combining vision-language reasoning with force-aware control could bridge perception and physical interaction for more robust real-world manipulation.
\section{Conclusion}
In this work, we introduce \model, a lightweight vision-language policy designed to address the challenges of inference efficiency, domain adaptation, and memory-based planning in robotic manipulation tasks. By leveraging a pre-trained 1B-parameter VLM and fine-tuning it on conversation-style robotic datasets, \model achieves state-of-the-art performance on VIMA-Bench while maintaining superior inference speed and data efficiency.

Our proposed multi-observation compression module significantly reduces the number of image tokens, leading to 47.5\% lower training time and a 34.1\% improvement in inference speed without compromising task success rates. Moreover, \model integrates historical memory and future goal images, enabling more effective long-horizon manipulation. Extensive evaluations demonstrate that \model outperforms the best-performing baseline by 18.8\% while operating under significantly lower computational constraints.

These results highlight \model as a promising step toward fast, memory-based and data-efficient vision-language policy generation models for robotic manipulation, paving the way for the integration of compact multi-modal models into real-world robotic applications.

{\small
\bibliographystyle{unsrt}
\bibliography{egbib}
}
\newpage

\end{document}